# Vehicle detection from GSV imagery: Predicting travel behaviour for cycling and motorcycling using Computer Vision

Kyriaki [Kelly] Kokka[1], Rahul Goel[2], Ali Abbas[1], Kerry A. Nice[3], Luca Martial[1], SM Labib[4], Rihuan Ke[5], Carola Bibiane Schönlieb[6], James Woodcock[1]

1. MRC Epidemiology Unit, University of Cambridge, Cambridge, UK
2. Transportation Research and Injury Prevention Centre, Indian Institute of Technology Delhi, New Delhi, India
3. Transport, Health, and Urban Systems Research Lab, Faculty of Architecture, Building, and Planning, University of Melbourne, Parkville, VIC, Australia.
4. Department of Human Geography and Spatial Planning, Utrecht University, Netherlands
5. School of Mathematics, University of Bristol, Bristol, BS8 1QU, United Kingdom
6. Department of Applied Mathematics and Theoretical Physics (DAMPT), University of Cambridge, Cambridge CB3 0WA, United Kingdom

## Abstract

**Introduction**

Transportation influence health by shaping exposure to physical activity, air pollution and injury risk. Comparative data on cycling and motorcycling behaviours is scarce, particularly at a global scale. Street view imagery, such as Google Street View (GSV), combined with computer vision, is a valuable resource for efficiently capturing travel behaviour data. This study demonstrates a novel approach using deep learning on street view images to estimate cycling and motorcycling levels across diverse cities worldwide.

**Methods**

We utilized data from 185 global cities. The data on mode shares of cycling and motorcycling estimated using travel surveys or censuses. We used GSV images to detect cycles and motorcycles in sampled locations, using 8000 images per city. The YOLOv4 model, fine-tuned using images from six cities, achieved a mean average precision of 89% for detecting cycles and motorcycles in GSV images. A global prediction model was developed using beta regression with city-level mode shares as outcome, with log transformed explanatory variables of counts of GSV-detected images with cycles and motorcycles, while controlling for population density.

**Results**

We found strong correlations between GSV motorcycle counts and motorcycle mode share (0.78) and moderate correlations between GSV cycle counts and cycling mode share (0.51). Beta regression models predicted mode shares with $R^2$ values of 0.614 for cycling and 0.612 for motorcycling, achieving median absolute errors (MDAE) of 1.3% and 1.4%, respectively. Scatterplots demonstrated consistent prediction accuracy, though cities like Utrecht and Cali were outliers. The model was applied to 60 cities globally for which we didn't have recent mode share data. We provided estimates for some cities in the Middle East, Latin America and East Asia.

**Conclusion**

GSV images offer valuable insights for predicting modes of cycling and motorcycling worldwide. GSV's capacity to discern travel modes and document street-level activity with the use of computer vision, makes it a complementary data source alongside the traditional methods.

# Introduction

Data on how people move as they go about their daily lives in cities is critical for research in transportation, public health, road safety, climate change, and for designing effective policies and interventions. For health, transport choices shape the exposure to physical activity, air pollution, and injury risk. Global trends show rising levels of physical inactivity (Strain et al., 2022), with a widespread shift toward private motorized transport. For two-wheeled modes, cycling and motorcycling represent two contrasting paths with different influences for public health. However, there is a lack of comprehensive data on cycling and motorcycling behaviours, and when studies assessing these modes are usually limited to a small number of cities. (Dill and Gliebe, 2008; Chen et. Al, 2022; Ngoc et al., 2022).

Cycling is a healthy, zero-emission mode of transport. It offers an effective way to incorporate physical activity into sedentary lifestyles. Regular physical activity reduces the risk of non-communicable diseases, poor physical and cognitive function, and support mental well-being (Bull et al, 2020; Wu et al., 2021; Strain et al., 2022). As such, cycling is receiving international interest as a sustainable and health-promoting transport mode (Mueller et al., 2015). However, injury risks are often high due to lack of protection from motor traffic. In contrast, motorcycling, though popular and growing rapidly in many low- and middle-income countries (LMICs) due to its affordability and ability to navigate congested streets, offers no physical activity benefits. Instead, it is associated with increased risks of traffic injuries both for users and pedestrians (WHO, 2022) and contributes to traffic related air pollution (Heydari et al., 2022).

We need consistent and up to date data on use of cycling to estimate trends of time, levels of physical activity, and injury risks (measured per bn km). Similarly we need data on motorcycling to estimate contribution to traffic related air pollution and greenhouse gas emissions and injury risks.

There are many ways to quantify such travel behaviours in cities and the most common is through mode share, defined as the percentage of trips by a specific mode of transport. Mode share is traditionally measured using household travel surveys that are conducted at city, regional, or national levels (Goel R et al., 2021). However, city-level travel surveys are uncommon across low-and-middle income countries. National travel surveys, that are often conducted in high-income countries (Goel et al., 2021) are often underpowered for many cities. Census data are only available in some settings and focus on commuting. Moreover, these surveys are time and resource consuming, and therefore not always conducted regularly. They capture passenger travel and not vehicles and the populations surveyed is of people living in the area and not travelling through it. Methods also vary between settings. In short, there is no globally consistent and comparable source with notable gaps in parts of the world. There is a need for more efficient ways in which researchers frequently capture how people move as they go about their daily lives worldwide.

The use of street view images is a potential data source for measuring travel behaviours. Global map service providers such as Google Street view (GSV) and Mapillary, or locally collected images from street cameras have been used in many studies to mainly detect characteristics of the built environment that determine behaviours related to transport (Doiron et al., 2022; Junehyung et al., 2023; Jiang et al., 2022; Biljecki and Koichi, 2021) and to lesser extent behaviours (Bai et al., 2023; Koo et al., 2022; Lu et al., 2018). However, to enable research with street images at scale and across contexts, the use of automated methods for extraction of relevant data from visual information is needed (Sánchez and Labib, 2024).

Computer vision (CV) is a branch of computer science focused on developing methods to interpret visual data akin to human vision, primarily leveraging deep learning models. These models have transformed image processing by enabling automated feature extraction and object recognition. CV models excel in tasks such as image classification, object detection, and semantic segmentation (Szeliski, 2022). They are adept at extracting various characteristics of the built environment with high accuracy and have been extensively employed in analysing street images to identify objects like cars, traffic signs, and trees (Yuhao et al., 2020), as well as behaviours like seatbelt usage and helmet-wearing among motorcycle drivers (Merali et al., 2020) or various two-wheeler helmets (Jakubec et al., 2023).

Considering the gaps in approaches to collect mode share data in diverse urban contexts, in this study a novel approach combining GSV images and deep learning is proposed to estimate levels of cycling and motorcycling for cities worldwide. This methodology represents the first attempt to leverage extensive GSV dataset with deep learning algorithms for global street-level behaviour assessment. It expands the previous proof of concept work that used manual annotations of GSV images to calculate mode shares of 7 different modes of transport and predicted mode shares for those in the cities of Great Britain (Goel et el., 2018). In the current study, we will deploy a deep learning model to automate travel behaviour data extraction in 185 cities around the world (including both middle- and high-income countries). The aim of this study is to investigate if the use of street imagery could predict use of travel modes for cycling and motorcycling in various contexts around the world and create a global prediction model using GSV observations and other data to make predictions.

## Methods

We used different sources of data and combined them in a beta regression model to create a global prediction model of city-level mode shares of cycles and motorcycles and applied the model in cities that did not have mode share information to predict their mode shares. Supplementary material 1 shows the flowchart of the methods used in this study.

### Study area

We gathered data from 185 cities located in LMICs and high-income countries (HIC). We used cities where all data (images and mode shares) were available for training and applied the trained model in cities with images but no available mode share. We refer to the two sets of cities as training and demo cities. The training cities were chosen based on having data for cycling and motorcycle mode shares, extensive GSV coverage and a population of roughly over 100,000 people. Similarly demo cities were selected if they had GSV coverage and population approximately over 100,000 people. This selection aimed to create a comprehensive sample of cities covering large part of the world, ensuring the representation of a range of cycling and motorcycling levels. The city boundaries and road networks were obtained from OpenStreetMap.

### Data sources

Our research drew from a unique dataset combining information on cycling and motorcycling mode shares, population density and GSV.

Cycling and motorcycling mode shares

The starting point for our analysis was the data collation on cycling and motorcycling use. This involved collating mode shares (in percentages) from a wide range of secondary data sources such as published research papers, city-level reports, or travel surveys. These sources had used either household travel surveys or census data. We collected mode shares at the city level from European countries (Belgium, Bulgaria, Czech Republic, UK, Finland, France, Germany, Hungary, Ireland, Netherlands, Norway, Poland, Romania, Serbia, Spain, Sweeden, Ukraine, and Italy), Asia (Hong Kong, Japan, Thailand, Turkey, Indonesia, Bangladesh, Philippines, Taiwan, India, Israel), Africa (Uganda, South Africa and Kenya), South America (Argentina, Brazil, Chile, Colombia, and Mexico), and North America (Canada and USA). There was a variation of commuting and all-trips purpose surveys. The date range was between 2011 and 2020; the majority was between 2016 and 2020. Other sources included KiM for Netherlands, MiD2017 for Germany, Epomm data, Eurostats for Italy, Census for the UK(adjusted to overall trips), travel surveys (e.g. Belgium, Norway and France), academic papers, and reports. The variation in the dates comes from the frequency of these surveys. The Census for the UK is conducted every 10 years and focuses on commuting trips. ADETEC (France) covers all-purpose trips periodically. Norway's National Travel Survey (RVU) covers all-purpose trips every 4-5 years. KiM (Netherlands) annual surveys cover all-purpose trips, while Germany's MiD is about all-purpose trips, conducted every 5 years. For the cities in the USA, mode shares of cycles and motorcycles was available from Census, which only reported trips on commuting to work. Using earlier work on cycling across a global set of cities (Goel et al., 2018), had reported a scaling factor to convert commuting mode shares to overall mode shares. We adjusted the mode shares for American cities by multiplying them by 0.72 due to their exclusive focus on commuting trips (Goel et al., 2018). We also adjusted the UK mode shares for all purpose trips. The dataset along with information including sources, areas and dates are available online: http://computervision.shinyapps.io/mode_share_atlas .

Google Street view

Key predictors of mode share were constructed using GSV images for geocoordinates of sampled locations using the Python OSMnx (Boeing, 2017) package with OpenStreetMap. The sample points were selected on the road network if a GSV image was available for the point (by checking the metadata). The sampling of locations was conducted to ensure a distance of 20m to 100m; depending on the GSV coverage and road density of each city. Next, we used GSV API (citation) to access the images. Input parameters for the API included the following: image size (640 × 640 pixels), geographic location (geographic coordinates), field of view (zoom level), up or down angle of the camera relative to the vehicle (default is 0), and heading (direction the camera is facing, with 0 = north, 90 = east, 180 = south, and 270 = west) along with the API key. We obtained 4 GSV images (directions: 0, 90, 180,270) for each location to capture 360-degree views of the environment. In total, 1,480,000 images were collected (2000 locations/8000 images per city for 185 cities). The images were acquired with varying dates because the API returns the latest image available for every location. Based on the metadata from 13 cities, images were captured from 2008 to 2022, with most of them taken after 2018 (see supplementary material 2). Manual review of images from 10 cities revealed issues with very few images, such as night-time scenes, indoor shots, and blurred images. These images were not replaced, and noise from such images was not removed due to their limited number (fewer than 5 locations per city).

Dates

We used available dates from mode shares and GSV images. For the mode shares, we recorded the year of the travel survey. If the travel survey was conducted over multiple years, we took the average year as the representative date. For the images, we recorded the year each image was collected. Thus, for each city, we had two dates corresponding to the cycle and motorcycle mode shares and 8000 dates for the images.

Population density

For geospatial information on population, we used the Global human settlement dataset (Schiavina M, 2021). This open access dataset provides data on the number of people in an area at a detailed resolution (250mx250m grids). We calculated the population density based on the total population counts within the city boundaries divided by the total area size (in km$^2$). We accounted for population density as it is an important factor for travel patterns (Goel R., 2020) and because in denser cities one would expect more activity of all types so we expect it to affect GSV counts.

YOLOv4 object detection model: Extracting travel behaviours from GSV images

The set of explanatory variables comprises the total counts of cycles and motorcycles in GSV images for every city. We refer to the GSV counts as **GSV cycle** and **GSV motorcycle**. These vehicle counts were detected using the YOLOv4 (Bochkovskiy A., 2020) object detection algorithm, chosen for its efficacy and efficiency in identifying vehicles. To enhance accuracy and generalizability across diverse urban landscapes, we fine-tuned the YOLOv4 model, pre trained in COCO (Lin et al., 2014) dataset, specifically for cycles, motorcycles, cargo cycles and e-rickshaws, incorporating images randomly selected from six cities (Rome, Bangkok, Kampala, Tokyo, Bogota, and Tel Aviv) to capture real-world complexities. Through *Labeling* software (Tzutalin, 2015), we labelled a total of 2713 images across four classes, achieving an average precision (AP) of 87% for motors, 91% for pedals, 83% for cargos, and 93% for e-rickshaws with confidence and intersection over union thresholds of 0.25 and 0.50 respectively, resulting in a mean average precision with the overlap of the predicted and the ground truth bounding box ≥ 0.50 (mAP@0.50) of 89%, recall of 0.78, precision of 0.87, and an f1 score (= $2 \times \frac{precision + recall}{precision \times recall}$) of 0.83. We included cargo cycles and e-rickshaws to make sure there will be no false positives but didn't consider their GSV counts in the prediction model. More details on the performance of the model and the training dataset are in the supplementary material 3-7.

**Global prediction model based on Beta regression**

We employed Pearson's correlation matrix to inform variable selection for the regression model. Subsequently, we utilized beta regression models to develop a prediction model for cycling and motorcycling mode shares. Beta regression models (Ferrari, S., & Cribari-Neto F., 2004) tailored for continuous variables ranging from 0 to 1, are suitable for modelling mode shares. We used betareg R library to implement beta regression for cycling and motorcycling mode shares. We log-transformed the explanatory variables to reduce their strong right-skewed distributions (see Supplementary material 8).

We developed two models, one for cycle mode share and one for motorcycle mode share using the same explanatory variables as shown in equation (1).

(1) $mode\ share \sim log(GSV\ cycle) + log(GSV\ motorcycle) + log(population\ density)$

To investigate if the difference in dates has any impact, we run weighted beta regression model for both modes using as weights (equation 2) the inversed difference between the mode share date and the date of every image.

$$(2)\ weights = \sum_{n=1}^{8000} \frac{1}{|mode\ share\ date - GSV\ image\ date_n|}$$

To evaluate model performance, we conducted leave-one-out cross-validation (LOOCV) to ensure robustness and unbiased assessment by iteratively excluding one observation and training the model with the remaining data. Root mean squared error, mean absolute error (MAE), median absolute error (MDAE), and R squared were computed using LOOCV (leave-one-out cross-validation). Additionally, Akaike information criterion (AIC), Bayesian information criterion (BIC), and log-likelihood (logLik) were calculated to compare model performance consistently across different types while considering data, variables, and sample size. Lower AIC and BIC values with higher logLik are preferred indicators of model effectiveness.

## Results

Based on figure 1, the correlations ranges from 0.42 to 0.78. GSV motorcycle has high correlation (0.78) with motorcycle mode share. Moderate correlations (0.51) exist between GSV cycle and cycling mode share. The correlation between GSV motorcycle and motorcycle mode share is 0.73. The correlations of GSV motorcycle with cycle mode share and GSV cycle with motorcycle mode share are – 0.01 and 0.32 respectively. In addition, GSV cycle and GSV motorcycle have moderate correlations with population density (0.55 and 0.64, respectively).

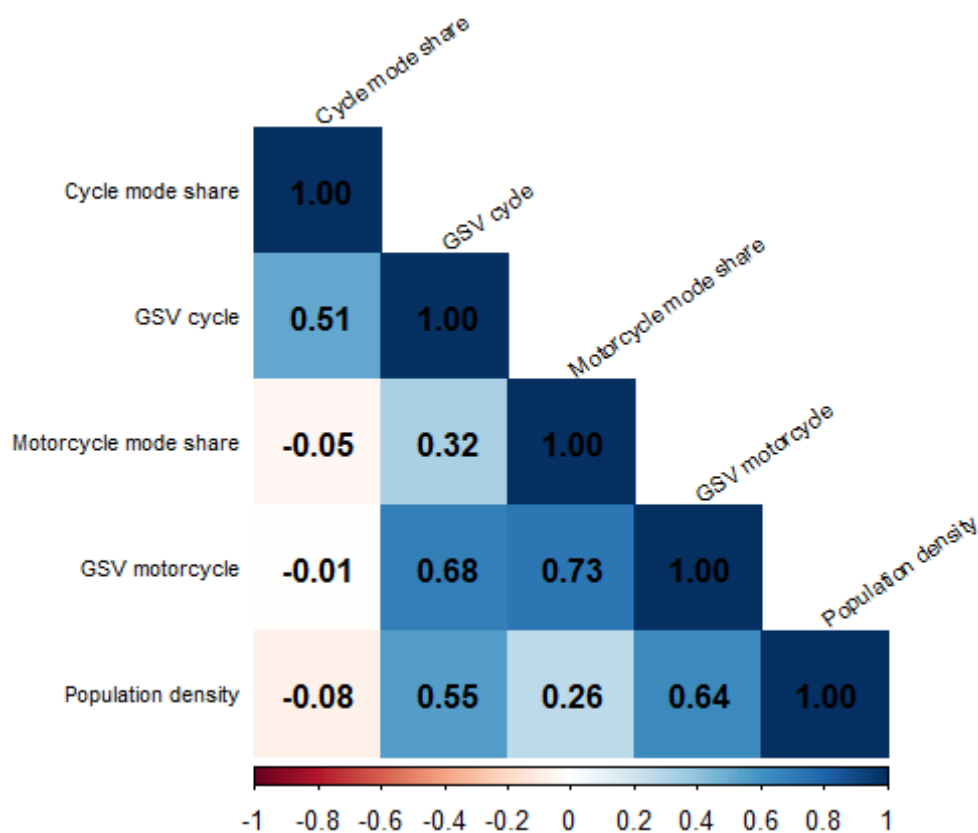

*Figure 1: Correlation matrix log transformed explanatory variables*

## Global prediction model: Beta regression

Table 1: Characteristics of the variables used in beta regression models

| description | variables | median | mean | Std.Dev. |
|---|---|---|---|---|
| **cycling mode share (n = 110)** | | | | |
| Transport indicators | GSV cycle | 155 | 285 | 396 |
| | GSV motorcycle | 142 | 435 | 717 |
| | Cycle mode share (%) | 3.1 | 6.1 | 8.4 |
| Density type indicator | Population density | 3502 | 4360 | 3421 |
| **motorcycling mode share (n = 95)** | | | | |
| Transport indicators | GSV cycle | 159 | 245 | 290 |
| | GSV motorcycle | 141 | 436 | 707 |
| | Motorcycle mode share (%) | 1 | 6.2 | 12 |
| Density type indicator | Population density | 3379 | 3976 | 2898 |

Table 1 shows descriptive summary of the indicators used in the regression model. For the cycling model We used all data points in the dataset. Table 2 presents the mean (MAE) and median of absolute errors. Figure 2 show the scatterplots of the observed and predicted values for cycling and motorcycling model based on the following equations:

$$cycle\ mode\ share = 1.138 * log(GSV\ cycle) - 0.39 * log(GSV\ motorcycle) - 0.863 * log(population\ density)$$

$$motorcycle\ mode\ share = -0.34 * log(GSV\ cycle) + 1.48 * log(GSV\ motorcycle) - 1.178 * log(population\ density)$$

Table 2: Metrics for beta regression models

| Beta regression | **Pseudo R squared** | AIC | BIC | RMSE | MAE (mean) | MDAE (median) |
|---|---|---|---|---|---|---|
| Cycling | 0.614 | -535 | -525 | 3.3 | 2.2 | 1.4 |
| Motorcycling | 0.612 | -531 | -518 | 5.1 | 2.9 | 1.3 |

The scatterplots of the observed and predicted values of mode shares (fig 2) show that the predicted values are scattered uniformly around and close to *y = x* line for both modes. This shows consistent accuracy of prediction across the full range of data for both modes. The MDAE are 1.4, and 1.3 for cycles and motorcycles respectively. The corresponding median predicted mode shares are 3.3 and 2.02, respectively (see table 2 and more details in Supplementary material 9-13).

The difference between MAE and MDAE is the highest for motorcycles. The highest difference between observed and predicted values is in Paris (2 and 17.96 respectively). Also in Cali, Colombia the model predicts a motorcycle mode share of 1.78%, which is an error of 14.82 percentage points as the observed value is 16.6%, thus substantially increasing the mean error. Additionally, Kaohsiung, Palembang, and Trieste had an error of more than 10 percentage points. In the scatterplot of motorcycles (Fig 3), cities with a predicted or observed share between 0 and 0.5 are not shown.

For cycling mode share two cities had a difference more that 10 pp between observed and predicted values. the highest difference between observed and predicted values is in Khulna, Bangladesh (5.9 and 20 respectively). In addition, Utrecht had an error of more than 10 percentage points. Overall, the max difference was 14.09 percentage points and the min 0.013 percentage points. We investigate if the error was due to the difference between the dates of the mode share and the images for these cities. Weighted beta regression didn't improve the model as the MAE is 5.58% and MDAE 3.89% for cycles and 8.28% and 5.37% for motorcycles respectively.

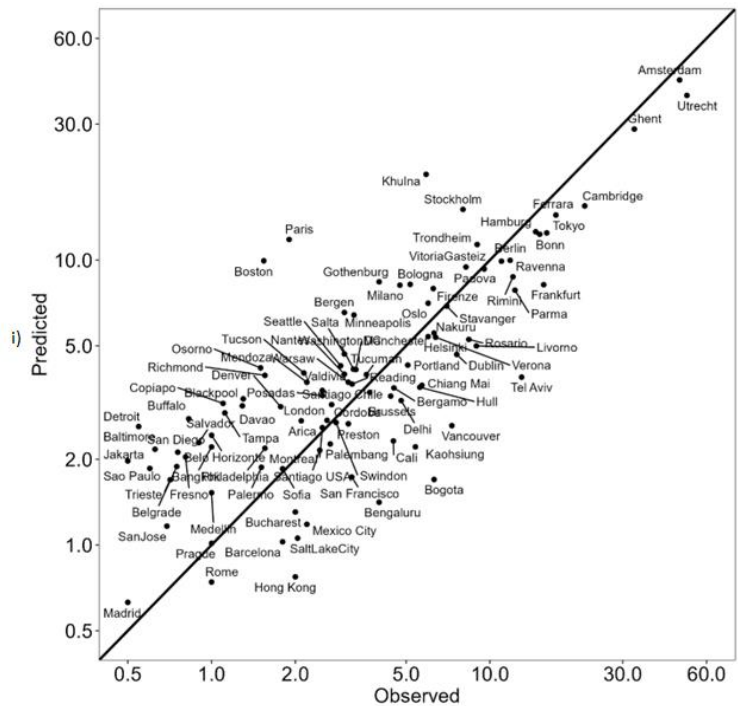

i)

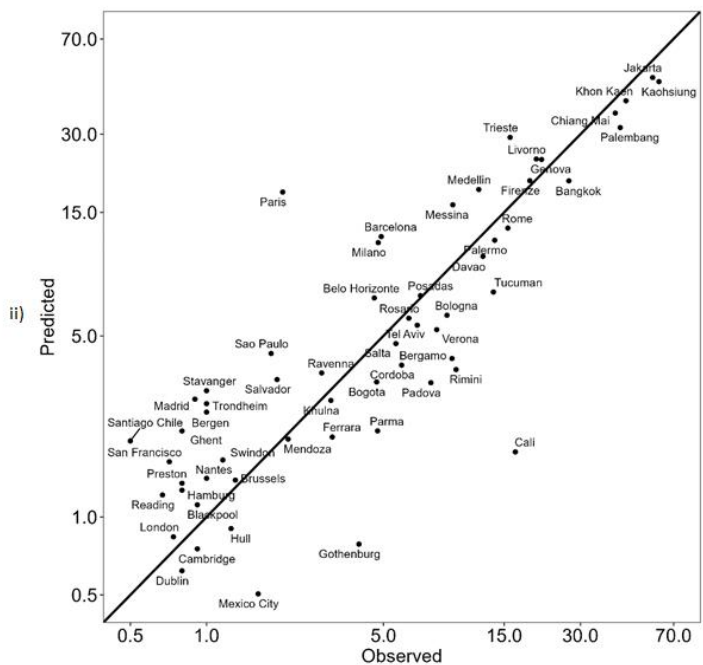

ii)

Figure 2: Observed vs predicted cycle(i) and motorcycle mode share(ii) (values between 0 and 0.5 are not visible in the scatterplot)

**Demo cities**

We applied the prediction model to estimate cycling and motorcycling mode shares in cities with GSV images but no mode share data. Overall, demo cities followed trends seen in training cities, with higher motorcycling shares in Asia and higher cycling shares in Europe. Among demo cities, cycling mode shares ranged from 1% (Hsinchu) to 43% (Copenhagen), with a median of 3% (mean = 6.9%), while motorcycling ranged from 0.37% (Rotterdam) to 65% (Hsinchu), with a median of 2% (mean = 6%). Cities with high cycling shares tended to have low motorcycling shares and vice versa, a pattern also seen in training cities.

Figure 3 visualizes the geographic distribution of mode shares, with training cities shown in green and demo cities in orange. Darker shades indicate higher mode shares. This highlights regional differences: across all cities, motorcycling was most prevalent in Asia (often more than 10%), while cycling was most common in Europe.

Among demo cities, cycling mode shares were highest in the Netherlands, Belgium, and across Scandinavia, ranging from 10% to 43%. In contrast, Eastern European countries showed lower cycling shares (2-4%) and very low motorcycling shares (below 1%). Latin and North America had modest cycling shares (1-4%). Southern European countries like Italy and Spain had higher motorcycling shares, with intra-country variation, e.g. Spanish cities ranged from ~2% in Leon to ~10% in Valencia.

In training cities, cycling was also highest in Europe, with variation both between and within countries. Eastern Europe had low cycling shares (1-3%), while Central Europe and Scandinavia had generally higher values. In Italy, mode shares ranged widely from 0.3% in Catania to 22% in Bolzano. In Spain, Madrid had 0.5% while Vitoria-Gasteiz reached 8.2%. Scandinavian cities ranged from 3% in Bergen to 23% in Umea. In the UK, extreme differences were between Blackpool reporting 1% and Cambridge 22%. In North America, cycling mode shares were mostly low, ranging from 0.2% in El Paso and Dallas to 7% in Portland. In South America, Sao Paulo recorded 0.6%, while Rosario reached 8.4%. Motorcycling was highest in Asia, with shares ranging from 12% to 61%. In Europe, it was generally low, especially in Eastern and Central Europe, the UK, and Scandinavia, where values were mostly below 1%. Italy was the exception, showing high motorcycling shares with within-country variation, from 3% in Reggio Emilia to 21% in Genova. In North America, motorcycling remained below 1%. South American cities varied: Colombian cities exceeded 10%, while Chilean cities remained under 1%.

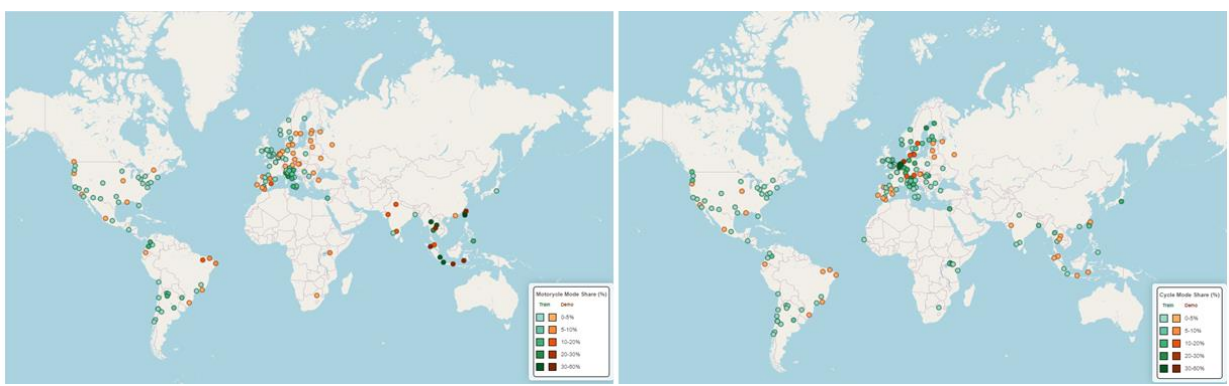

Figure 3: Geographic distribution of mode shares (training cities in green and demo cities in orange with darker shades indicate higher mode shares)

## Discussion

### Main findings

We investigated relationships between GSV-based observations of cycles and motorcycles using deep learning and city-level travel behaviour for 185 cities around the world. We found that GSV observations are good predictors of mode shares of cycling and motorcycling. We found correlation of R = 0.614 between log GSV counts cycle mode share. The correlation between the log GSV counts and motorcycling mode share was R = 0.612. Including mode share dates and image dates didn't improve the prediction models for neither cycling nor motorcycling. These findings emphasize the use of GSV imagery combined with CV methods in predicting travel behaviour patterns and highlights its potential as a valuable global tool for public health planning and policymaking.

### What might have influenced the differences between predicted and observed

GSV based estimates and travel surveys are likely capturing different but related latent variables. The motivation for this work is the limitation of travel survey data, in terms of coverage spatially and temporally and in terms of consistency. Thus our training data is not a gold standard data set but a messy real world one. Therefore when we observe inconsistencies between our predicted and reported mode shares it is not obvious which is 'correct'. For example the model overestimated motorcycle use in Paris, Cali, and Trieste.

While the deep learning model performed well, some external factors likely influenced cycle and motorcycle counts and consequently the prediction models. The accuracy of mode share predictions might vary across cities due to factors like occlusions, cargo loads, traffic density, and lighting conditions of the images. Despite this, evaluation of the CV model and manual check of the detections across 13 cities confirmed minimal misclassifications, mainly with cycling infrastructure signs, and not between bicycles and motorcycles. Timing of image collection may have led to underestimations, though detecting parked vehicles helped mitigate this.

### Compare and contrast with the literature

There are very few studies that used street images and automated data extraction to assess transport behaviours and to the best of our knowledge none accessed multiple global cities. A study by (Biljecki et Ito; 2021) assessed bikeability in Tokyo and Singapore by extracting several aspects of the built environment and created a bikeability index. Another study used seven Canadian cities and Census data to predict walking to work from street images (Doiron et al., 2022). They found that features derived from street images using CV are better to predict the percent of people walking to work as their main mode of transport compared to data derived from traditional walkability metrics. They detected people, buildings and sky from images and used them as predictors in the regression model.

### Strengths and limitations

Our study has several strengths. It is the first to investigate the use of street imagery to estimate global cycling and motorcycling mode shares at the city level, utilizing variables such as cycling and motorcycling counts from GSV images and population densities. This makes the prediction models applicable to any city with available street images. By applying CV methods to a vast GSV data collection from global cities, we automated the collation of predictors, showcasing the potential of these technologies to evaluate travel behaviours quickly and effectively. CV allowed us to upscale the study, fine-tune the model with global data, and label images from diverse settings, creating a model that generalizes well across diverse cities. Our open-source models can be replicated or used in other settings with available street images. We found that even a small sample size of 1000 images per city was sufficient for stable observations so future studies could lower their sampling points. GSV serves as a consistent and accessible data source across various settings. While GSV and travel diary-based mode share do not capture the same thing, GSV might say more specifically about the use of a mode in the city. Overall, our study enriched methods and datasets for predicting cycling and motorcycling mode shares worldwide using GSV and CV methods, offering consistent insights through large-scale analysis.

Some limitations of our study include the spatial and temporal constraints of GSV. Spatially, GSV lacks data on inaccessible local backroads and alleys and generally covers and frequently update mostly central urban areas (Rzotkiewicz et al., 2018). Temporally, GSV updates are globally slow and inconsistent, potentially leading to patches of images along the same road particularly for informal or secondary roads or cities in developing countries (Fry et al., 2020; Sánchez and Labib, 2024). The metadata reports only the month and year, not the time or day of the week, which are likely significantly associated with GSV observations and could introduce measurement error or bias. Furthermore, GSV data collection does not occur during rain, affecting active travel observations. Variations in the number of rainy days between cities could introduce bias, although this is potentially controllable if the underlying relationships are known. There also a cost associated with purchasing the images and access limitations (Helbich et al., 2024) although some countries laws provide exceptions for researchers to freely scrap data strictly for research use.

Traditional mode share data sources, such as census, travel surveys or reports, each come with their own limitations. Travel surveys do not consistently capture recreational cycling. They also capture travel by public and not other users such as couriers or commercial trips. The size and representative of the sample will vary. Censuses only capture commuting, cycling tends to be more common for commuting than for other purposes in lower cycling settings, although the relationship varies. Some data are from published reports for which the underlying survey method and date are not entirely clear.

**Future work**

Street images are a promising source for understanding the relationship between built environment features and behaviours that are relevant to public health. Their integration with deep learning offers a deeper understanding of the urban built environment that could be used across various scales, including location, street, and city to perform broader environmental audits. This enhanced understanding provides valuable support for policymaking. Future work could implement a segmentation model. This could enable not only the identification of vehicles but also the determination of their area proportion and the state of parked or moving cycles with cyclists (or motorcycles) and additional features of the built environment relevant to transport behaviour (e.g., transport infrastructure). This would also allow linking the behaviours with the characteristics of the environment that influence these behaviours. However, the annotation process for fine tuning image segmentation models, which requires outlining the shape of each object, is a time-consuming task. Finally, we could include more global cities as GSV has recently become available in more places such as Lagos, Nairobi, Benin city, Dakar, Beirut and Doha.

# Conclusion

Street images offer valuable insights for predicting modes of cycling and motorcycling worldwide and could be used as a complementary data source to traditional methods like travel surveys. They provide a rich, visual dataset that captures transportation behaviours and characteristics of the environment that influence these behaviours. By using computer vision techniques on street images from 185 global cities, we automatically identified two travel modes, bicycles and motorcycles and accurately predict their mode shares.

## Acknowledgments

JW, KK, AA and LM received funding for this work from the GLASST project through the European Research Council (ERC) under the Horizon 2020 research and innovation programme (Grant Agreement No 817754). This material reflects only the author's views, and the Commission is not liable for any use that may be made of the information contained therein.

KK was supported by the Economic and Social Research Council Doctoral Training Partnership (ESRC DTP) Cambridge (ES/P000738/1).

KAN was supported by Australian National Health and Medical Research Council (2019/GNT1194959; grant number GA80134)

We would like to thank Anna Goodman at the London School of Hygiene & Tropical Medicine for providing the mode share data for England. We would also like to thank Angelica Aviles-Rivero at the Department of Applied Mathematics and Theoretical Physics for her assistance with the computer vision model.

Patients or the public WERE NOT involved in the design, or conduct, or reporting, or dissemination plans of our research.

# Supplementary material

**Supplementary material 1:** Flowchart of the project

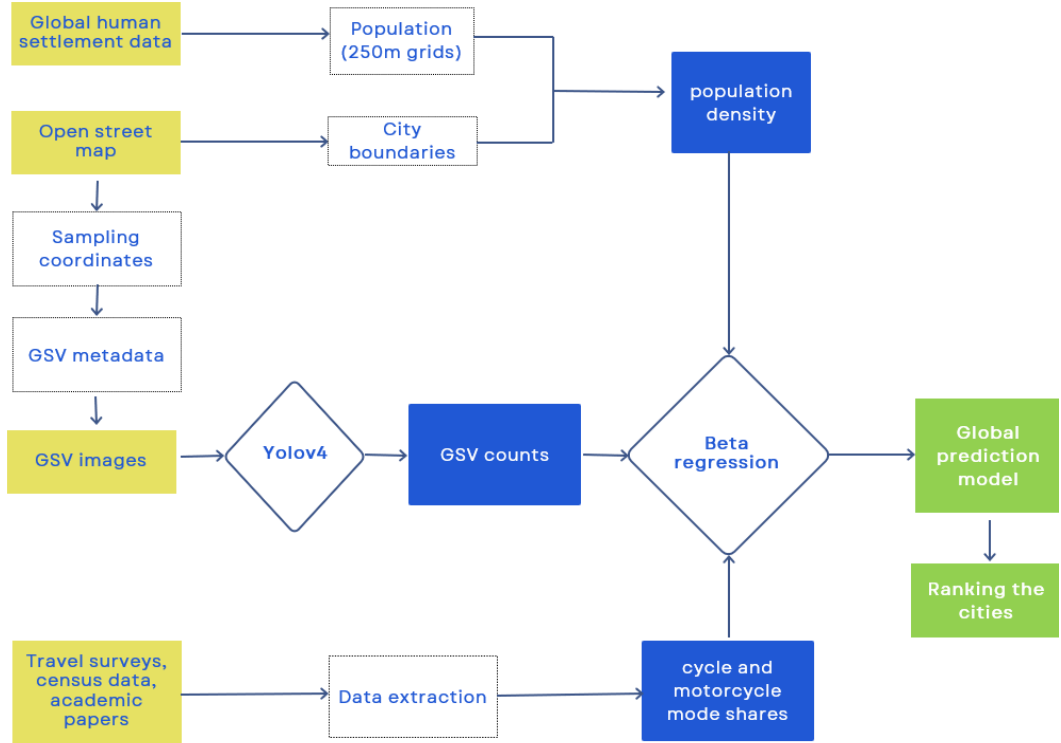

**Supplementary material 2:** GSV image dates for some cities

| Rosario | | Tucuman | | Cordoba | | Belo Horizonte | | Salvador | | Oslo | |
|---|---|---|---|---|---|---|---|---|---|---|---|
| year | | year | | year | | year | | year | | year | |
| 2008 | 1 | 2014 | 982 | 2013 | 424 | 2009 | 14 | 2011 | 14 | 2009 | 39 |
| 2013 | 519 | 2015 | 51 | 2014 | 170 | 2011 | 5 | 2012 | 30 | 2010 | 2 |
| 2014 | 1 | 2016 | 1 | 2015 | 93 | 2012 | 2 | 2013 | 24 | 2012 | 7 |
| 2015 | 20 | 2017 | 22 | 2016 | 1 | 2013 | 5 | 2014 | 1 | 2014 | 10 |
| 2016 | 2 | 2018 | 56 | 2017 | 33 | 2014 | 22 | 2015 | 43 | 2016 | 1 |
| 2017 | 50 | 2019 | 266 | 2018 | 108 | 2015 | 5 | 2016 | 15 | 2017 | 11 |
| 2018 | 89 | 2021 | 115 | 2019 | 790 | 2016 | 1 | 2017 | 23 | 2018 | 12 |
| 2019 | 819 | 2022 | 507 | 2020 | 105 | 2017 | 7 | 2018 | 272 | 2019 | 390 |
| 2020 | 225 | | | 2021 | 276 | 2018 | 189 | 2019 | 605 | 2020 | 409 |
| 2021 | 274 | | | | | 2019 | 507 | 2020 | 7 | 2021 | 78 |
| | | | | | | 2020 | 69 | 2021 | 302 | 2022 | 1041 |
| | | | | | | 2021 | 517 | 2022 | 664 | | |
| | | | | | | 2022 | 657 | | | | |

| Arica | | Cali | | Rimini | | Utrecht | | Cambridge | | Genova | |
|---|---|---|---|---|---|---|---|---|---|---|---|
| year | | year | | year | | year | | year | | year | |
| 2012 | 1313 | 2013 | 111 | 2010 | 16 | 2009 | 45 | 2008 | 18 | 2008 | 22 |
| 2014 | 685 | 2014 | 17 | 2012 | 11 | 2010 | 2 | 2010 | 2 | 2009 | 3 |
| 2017 | 1 | 2015 | 32 | 2014 | 9 | 2011 | 2 | 2011 | 2 | 2010 | 8 |
| 2020 | 1 | 2016 | 10 | 2015 | 4 | 2013 | 2 | 2012 | 92 | 2011 | 1 |
| | | 2017 | 28 | 2016 | 7 | 2014 | 58 | 2014 | 331 | 2012 | 47 |
| | | 2018 | 182 | 2017 | 13 | 2015 | 51 | 2015 | 87 | 2014 | 23 |
| | | 2019 | 1608 | 2018 | 15 | 2016 | 55 | 2016 | 44 | 2015 | 76 |
| | | 2020 | 2 | 2019 | 6 | 2017 | 72 | 2017 | 52 | 2016 | 10 |
| | | 2021 | 10 | 2020 | 54 | 2018 | 127 | 2018 | 156 | 2017 | 13 |
| | | | | 2021 | 3 | 2019 | 60 | 2019 | 176 | 2018 | 43 |
| | | | | 2022 | 1862 | 2020 | 378 | 2020 | 295 | 2019 | 41 |
| | | | | | | 2021 | 820 | 2021 | 740 | 2020 | 693 |
| | | | | | | 2022 | 330 | 2022 | 5 | 2021 | 253 |
| | | | | | | | | | | 2022 | 767 |

| Davao | |
|---|---|
| year | |
| 2015 | 182 |
| 2016 | 37 |
| 2017 | 7 |
| 2018 | 606 |
| 2019 | 1 |
| 2021 | 1167 |

| cities | Images collected <=2018 | Images > 2018 | Cities | Images collected <=2018 | Images > 2018 |
|---|---|---|---|---|---|
| Rosario | 682 | 1318 | Cali | 380 | 1620 |
| Tucuman | 1112 | 888 | Rimini | 75 | 1925 |
| Cordoba | 829 | 1171 | Utrecht | 412 | 1588 |
| Belo Horizonte | 250 | 1750 | Cambridge, UK | 784 | 1216 |
| Salvador | 422 | 1578 | Genova | 246 | 1754 |
| Oslo | 82 | 1918 | Davao | 832 | 1168 |
| Arica | 1999 | 1 | | | |

**Supplementary material 3:** Number of images per class for train/test for YOLOv4

| Classes | Training sample | Testing sample |
|---|---|---|
| Motorcycle | 1190 | 159 |
| bicycle | 673 | 129 |
| Cargo bikes | 86 | 24 |
| e-rickshaw | 385 | 67 |
| total | 2334 | 379 |

**Supplementary material 4**

Random sampling of 8000 images per city: We checked ratio of pedal (and motor) detections / sampling images for few cities. As an example in Baltimore we run 4 times with random locations selected each time and after 4000 images the ratio was plateaued. So 8000 randomly selected images per city are more than enough for the study.

Baltimore

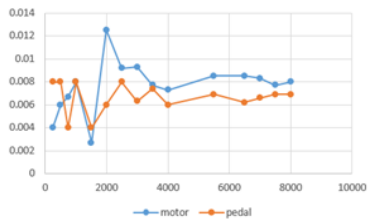
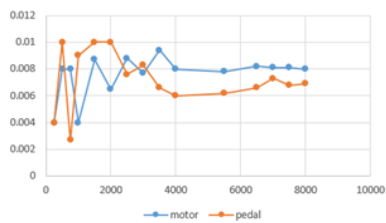
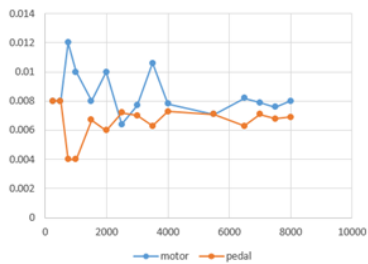
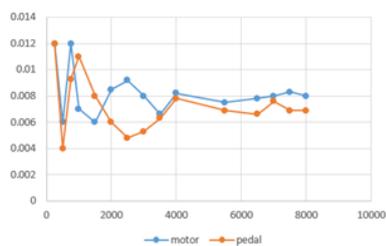

**Supplementary material 5**: Plots with YOLOv4 detected vs manually detected cycles, motorcycles and both vehicles combined

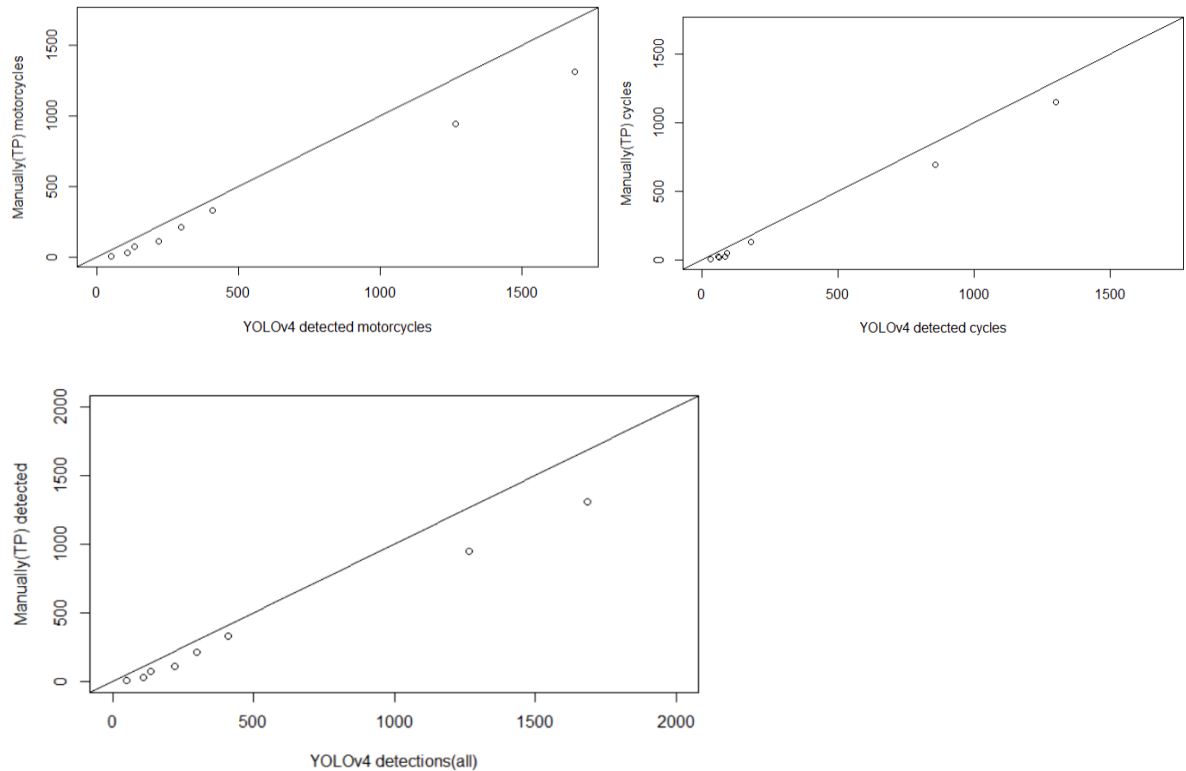

**Supplementary material 6:** Table with YOLOv4 detected vs manually detected cycles, motorcycles and both combined (same as the above plots)

| city | Cycle_yolo | Motorcycle_yolo | TP cycle | TP motorcycle | Yolo sum | TP sum |
|---|---|---|---|---|---|---|
| San Francisco | 117 | 181 | 86 | 127 | 298 | 213 |
| Boston | 133 | 86 | 88 | 21 | 219 | 109 |
| El Paso | 17 | 33 | 2 | 3 | 50 | 5 |
| Bonn | 317 | 92 | 285 | 46 | 409 | 331 |
| Hamburg | 359 | 110 | 323 | Na | 469 | Na |
| Kampala | 387 | 1299 | 161 | 1148 | 1686 | 1309 |
| Bogota | 408 | 857 | 255 | 689 | 1265 | 944 |
| Buffalo | 42 | 65 | 15 | 14 | 107 | 29 |
| Portland | 74 | 61 | 49 | 21 | 135 | 70 |

TP means manually true positive detection (based on manual counting of the objects)

**Supplementary material 7:** YOLOv4 metrics

confidence threshold: 0.25
IoU threshold: 0.50

| class | Average Precision (%) | TP | FP |
|---|---|---|---|
| motor | 87 | 131 | 39 |
| Pedal | 91 | 120 | 30 |
| Cargo | 83 | 20 | 7 |
| Rickshaw | 93 | 59 | 15 |

| Recall | Precision | F1 | mAP @0.50 |
|---|---|---|---|
| 0.78 | 0.87 | 0.83 | 89 % |

| Total TP | Total FP | Total FN |
|---|---|---|
| 330 | 91 | 50 |

**Supplementary material 8:** Histograms and log transformed histograms for all variables used in beta regression

| Explanatory variable | histogram | Log transformed histogram |
|---|---|---|
| **Cycling model** | | |
| GSV cycle | Histogram of c$pedal | Histogram of log(c$pedal) |
| GSV motorcycle | Histogram of c$motor | Histogram of log(c$motor) |
| Population density | 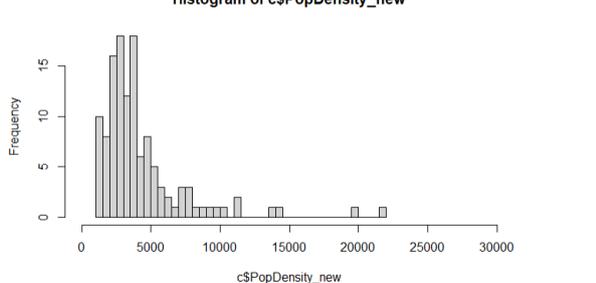 | 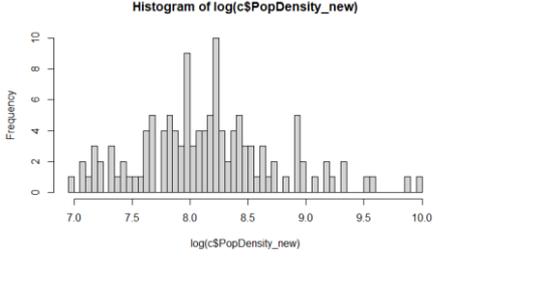 |
| Cycle mode share | 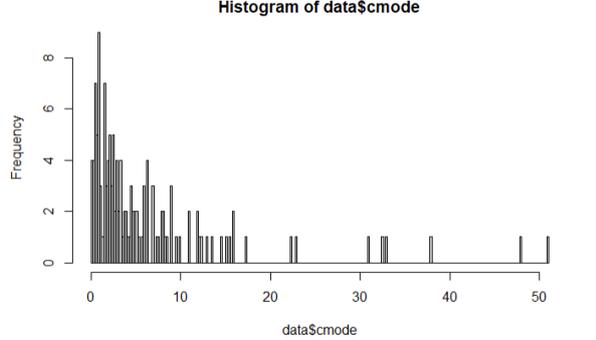 | |
| **Motorcycle model** | | |
| GSV cycle | 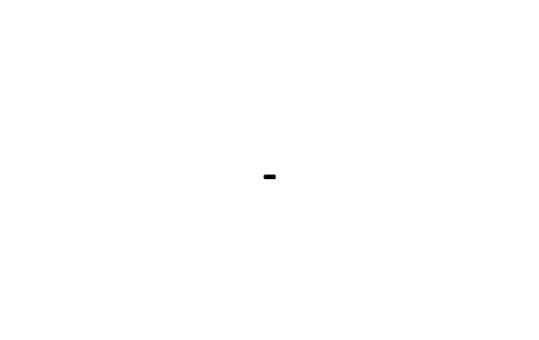 | 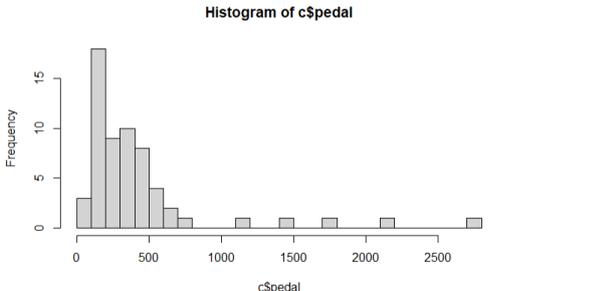 |
| GSV motorcycle | 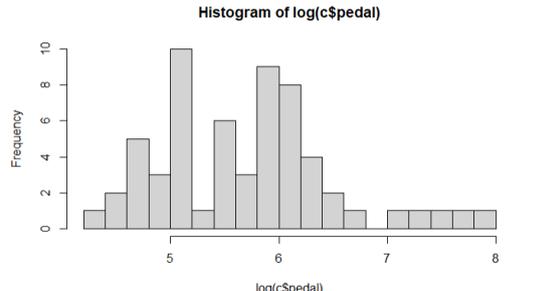 | 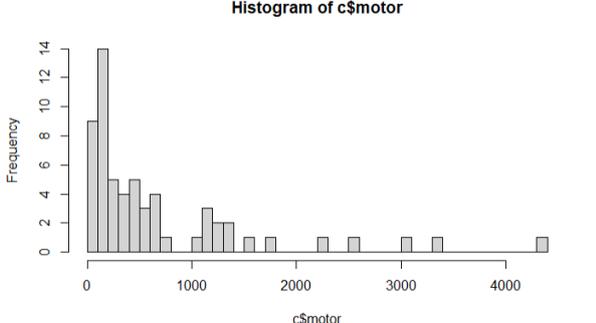 |
| Population density | 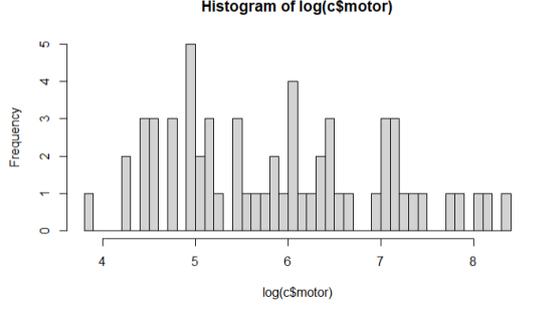 | 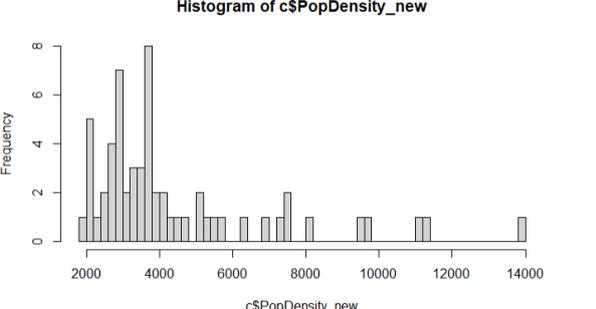 |

| motorcycle mode share | 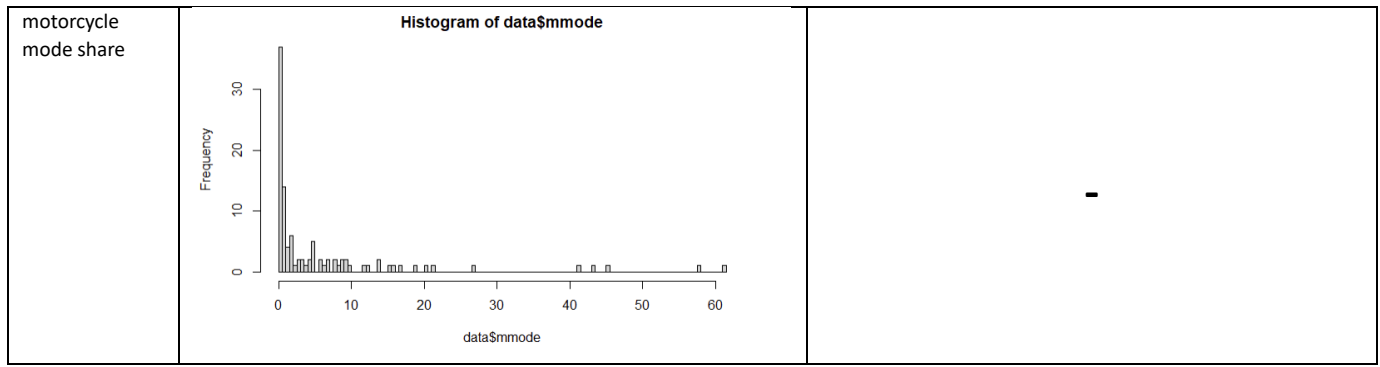 | - |

**Supplementary material 9:** Characteristics of training cities for cycling

| Variable | N | Mean | Std. Dev. | Min | Pctl. 25 | Pctl. 75 | Max |
|---|---|---|---|---|---|---|---|
| cmode | 110 | 5.4 | 8.2 | 0.05 | 1.1 | 6.2 | 51 |
| pedal | 110 | 285 | 396 | 17 | 81 | 349 | 2719 |
| motor | 110 | 435 | 717 | 32 | 72 | 456 | 4400 |
| PopDensity_new | 110 | 4360 | 3421 | 1084 | 2466 | 4780 | 21635 |
| ID | 110 | 56 | 32 | 1 | 28 | 83 | 110 |
| percentagecmode | 110 | 0.054 | 0.082 | 0.0005 | 0.011 | 0.062 | 0.51 |

**Supplementary material 10:** Characteristics of training cities for motorcycling

| Variable | N | Mean | Std. Dev. | Min | Pctl. 25 | Pctl. 75 | Max |
|---|---|---|---|---|---|---|---|
| pedal | 95 | 245 | 290 | 17 | 81 | 335 | 2163 |
| mmode | 95 | 6.2 | 12 | 0.01 | 0.27 | 6.5 | 61 |
| motor | 95 | 436 | 707 | 33 | 72 | 486 | 4400 |
| PopDensity_new | 95 | 3976 | 2898 | 1084 | 2194 | 4632 | 19694 |
| ID | 95 | 48 | 28 | 1 | 24 | 72 | 95 |
| percentagemmode | 95 | 0.062 | 0.12 | 0.0001 | 0.0027 | 0.066 | 0.61 |

**Supplementary material 11:** Regression model results

| | n | R squared | AIC | BIC | RMSE | MAE (mean) | MDAE (median) |
|---|---|---|---|---|---|---|---|
| Cycling(with intercept)beta | 110 | 0.635 | -521 | -507 | 4.1 | 2.5 | 1.5 |
| Cycling (without intercept)beta | 110 | 0.637 | -535 | -525 | 3.3 | 2.2 | 1.4 |
| Cycling beta without intercept date weights | 53 | 0.743 | -72 | -64 | 7.6 | 5.6 | 3.9 |

| Motorcycling (with intercept) beta | 95 | 0.612 | -531 | -518 | 5.1 | 2.9 | 1.3 |
| Motorcycling (without intercept) | 95 | 0.569 | -528 | -517 | 4.7 | 2.8 | 1.5 |
| Motorcycling beta without intercepts date weights | 33 | 0.916 | -130 | -124 | 10.8 | 8.2 | 5.4 |

**Supplementary material 12:** Min, median, mean and max for the predicted mode shares for cycling and motorcycling in demo cities

| Predicted mode share | min | median | mean | max |
| --- | --- | --- | --- | --- |
| Cycle | 0.66 | 3.43 | 5.53 | 43.15 |
| motorcycle | 0.37 | 2.02 | 6.35 | 49.66 |

**Supplementary material 13**: Cities with difference between observed and predicted motorcycle mode share > 10. For the difference min: 0.07, median: 1.45, mean: 2.57, max 15.96

| city | Observed motorcycle mode share | Predicted | dif |
| --- | --- | --- | --- |
| Paris | 2 | 17.96 | 15.96 |
| Cali, Colombia | 16.6 | 1.78 | 14.82 |
| Kaohsiung, Taiwan | 61.3 | 47.91 | 13.39 |
| Trieste | 15.83 | 29.19 | 13.36 |
| Palembang, Indonesia | 43.09 | 31.85 | 11.24 |